\documentclass[sigconf]{acmart}

\AtBeginDocument{%
  }

\copyrightyear{2024}
\acmYear{2024}
\setcopyright{acmlicensed}\acmConference[MM '24]{Proceedings of the 32nd ACM International Conference on Multimedia}{October 28-November 1, 2024}{Melbourne, VIC, Australia}
\acmBooktitle{Proceedings of the 32nd ACM International Conference on Multimedia (MM '24), October 28-November 1, 2024, Melbourne, VIC, Australia}
\acmDOI{10.1145/3664647.3681705}
\acmISBN{979-8-4007-0686-8/24/10}

\usepackage{colortbl}
\usepackage{bm}

\usepackage{xcolor}
\usepackage{threeparttable}
\usepackage{todonotes}
\usepackage{CJKutf8}
\usepackage{booktabs}
\usepackage{caption}
\usepackage{hyperref}  

\begin{document}

\title{Decoding Urban Industrial Complexity: Enhancing Knowledge-Driven Insights via  IndustryScopeGPT}

\author{Siqi Wang}
\affiliation{%
  \institution{Tongji University}
  \city{Shanghai}
  \country{China;}}

\affiliation{%
  \institution{Department of Computing, The Hong Kong Polytechnic University}
  \city{Hong Kong}
  \country{China}}
\email{siqi_wang@tongji.edu.cn}
\email{siqi23.wang@connect.polyu.hk}

\author{Chao Liang}
\affiliation{%
  \institution{Guangdong Guodi Institute of Resources and Environment}
  \city{Guangzhou}
  \country{China}}
\email{liangc0427@gmail.com}

\author{Yunfan Gao}
\affiliation{%
  \institution{Shanghai Research Institute for Intelligent Autonomous Systems, Tongji University}
  \city{Shanghai}
  \country{China}}
\email{2311821@tongji.edu.cn}

\author{Yang Liu}
\affiliation{%
  \institution{College of Design and Innovation, Tongji University}
  \city{Shanghai}
  \country{China}}
\email{liuy@tongji.edu.cn}

\author{Jing Li*}
\affiliation{%
  \institution{Department of Computing, The Hong Kong Polytechnic University;}
  \institution{Research Centre on Data Science \& Artificial Intelligence}
  \city{Hong Kong}
  \country{China}}
\email{jing-amelia.li@polyu.edu.hk}
  
\author{Haofen Wang*}
\affiliation{%
  \institution{College of Design and Innovation, Tongji University}
  \city{Shanghai}
  \country{China}}
\email{carter.whfcarter@gmail.com}

 \thanks{*Corresponding authors}

\renewcommand{\shortauthors}{Siqi Wang, et al.}

\begin{abstract}
Industrial parks are critical to urban economic growth. Yet, their development often encounters challenges stemming from imbalances between industrial requirements and urban services, underscoring the need for strategic planning and operations. This paper introduces IndustryScopeKG, a pioneering large-scale multi-modal, multi-level industrial park knowledge graph,  which integrates diverse urban data including street views, corporate, socio-economic, and geospatial information, capturing the complex relationships and semantics within industrial parks. Alongside this, we present the IndustryScopeGPT framework, which leverages Large Language Models (LLMs) with Monte Carlo Tree Search to enhance tool-augmented reasoning and decision-making in \textbf{I}ndustrial \textbf{P}ark \textbf{P}lanning and \textbf{O}peration (IPPO). Our work significantly improves site recommendation and functional planning, demonstrating the potential of combining LLMs with structured datasets to advance industrial park management. This approach sets a new benchmark for intelligent IPPO research and lays a robust foundation for advancing urban industrial development. The dataset and related code are available at \url{https://github.com/Tongji-KGLLM/IndustryScope}.
\end{abstract}




\maketitle

\section{Introduction}

Industrial parks are key engines driving economic growth and centers of innovation within cities. They connect the economy, living environments, and environmental sustainability, fostering the integration of technological innovation and urban life~\cite{circular_econmy}. However, many face a significant imbalance between industrial growth and urban service provision, leading to unsustainable development patterns~\cite{trends}. This imbalance highlights the urgent need for scientific planning and operation of industrial parks. Such operation requires a strategic consideration of local economic levels, infrastructure, and industrial foundations, aiming to provide optimization suggestion, public service facility site recommendation, and comprehensive industrial zone planning~\cite{optimization}. Traditional approaches, often based on empiricism and outdated surveys, fail to dynamically integrate rich urban data for deep analytical insights~\cite{eco-industrial-parks,park-develop-case}.

\begin{figure*}[thbp]
	\centering	\includegraphics[width=0.85\textwidth]{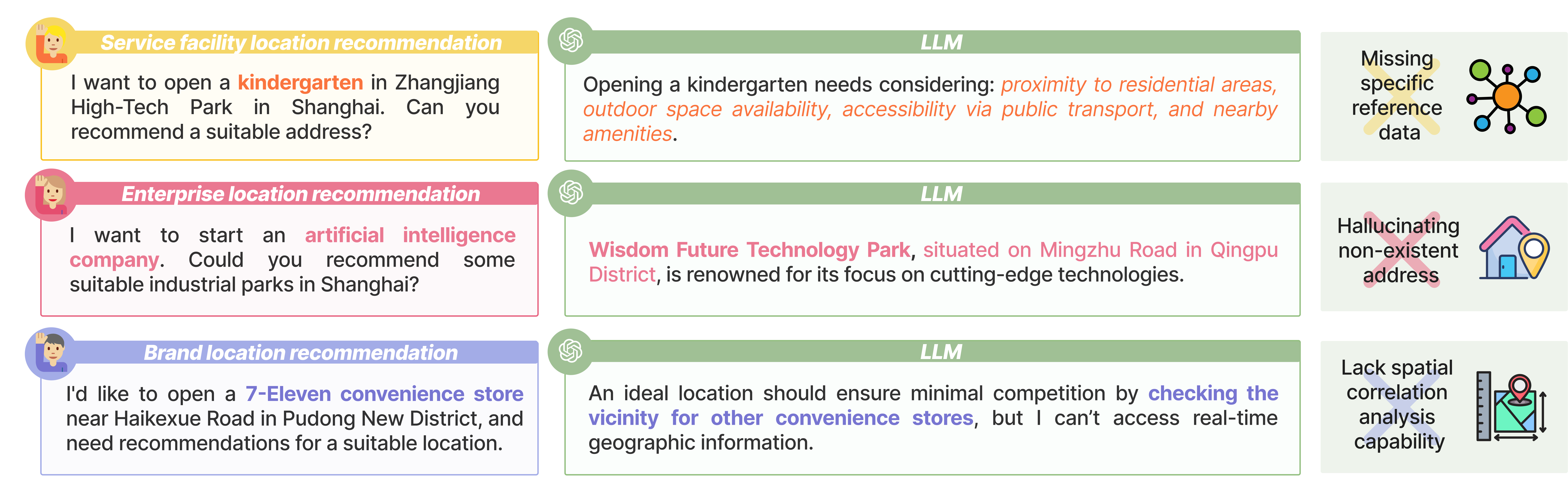}
	\caption{The challenges in integrating LLMs for IPPO solutions.}
	\label{fig1}
\end{figure*}

The advancement of artificial intelligence enhances urban tasks such as traffic management~\cite{traffic_prediction}, urban planning~\cite{Urban_Planning}, urban function prediction~\cite{function-prediction}, public safety~\cite{safety}, and site recommendation~\cite{OpenSiteRec}. However, intelligent industrial park planning and
operation (IPPO) remains largely untapped. The emergence of Large Language Models (LLMs)~\cite{chatgpt}, with their robust language reasoning and in-context learning capabilities~\cite{ICL}, offers new avenues for addressing urban complexities and developing unified, adaptive IPPO solutions.

{\bfseries Challenge 1: How to effectively construct an industrial park dataset capturing complex relationships and semantics?} Current datasets often overlook the detailed needs of industrial parks, focusing on geographical features and neglecting cultural and socio-economic aspects. Industrial park data is typically multi-source and heterogeneous, including information sources such as enterprises, government websites, statistical yearbooks, street views, etc. Although these data are not explicitly linked, they inherently share attributes and spatial relationships that constitute rich semantic information. To thoroughly evaluate an industrial park's development, it is essential to leverage multidimensional information. Knowledge graphs, by organizing and integrating these diverse data sources, offer a streamlined framework.

{\bfseries Challenges 2: How to adapt LLMs for industrial park KG?}  Industrial park knowledge graphs differ from general knowledge graphs by being heterogeneous, incorporating image, textual, numerical, and geospatial data with intricate entity relationships. Unlike LLMs focused on textual tasks, they require real-time geospatial data transmission and computations, necessitating graph databases and tools for live interactions (Figure~\ref{fig1}). To address the limitations of static knowledge in LLMs, integrating real-time knowledge graph databases and leveraging retrieval-augmented generation (RAG) methods~\cite{Rag_survey} offer a promising solution. Further research is needed to enhance LLMs' reasoning by bridging the gap between knowledge bases and user queries effectively.

{\bfseries Challenge 3: How can LLMs flexibly and interpretably excel in diverse IPPO tasks?} Traditional urban models, often trained on specific datasets for particular tasks, lack the necessary flexibility for broader applications. For example, the features considered for situating financial institutions versus restaurants differ significantly, making it costly to retrain models. Additionally, popular methods like spatial representation learning~\cite{CNN} can diminish the interpretability. Clear metrics and continuous reasoning are crucial for enhancing decision explainability in industrial park planning. While advancements in LLM multi-step reasoning~\cite{CoT} and agents~\cite{ReACT} show promise for flexibility and interpretability, these often operate in isolation and could benefit from integration with external graph databases to enhance reasoning capabilities.

To tackle the challenges mentioned, this paper introduces a pioneering work that constructs a multi-modal, multi-level large-scale industrial park knowledge graph, IndustryScopeKG. By extracting entities from diverse data sources and combining domain knowledge, a substantial industrial park knowledge base with various spatial and semantic relationships has been built. The IndustryScopeGPT framework is introduced to enable LLMs to dynamically adapt to the structure of knowledge graph and enhance decision-making capabilities through Monte Carlo Tree Search and reward information. The performance of the framework in IPPO tasks is validated through the development of IndustryScopeQA benchmark, demonstrating the reliability and advantages of the framework. Our contributions are summarized as follows:
\begin{itemize}
    \item We release the first open-source, multi-modal, multi-level (spatial and semantic level) large-scale knowledge graph dataset, IndustryScopeKG, for IPPO tasks.
    \item We introduce the IndustryScopeGPT framework, which enhances LLMs' planning, action, and reasoning capabilities through the integration of graph databases and various tools, along with Monte Carlo Tree Search for optimal reasoning paths. It represents the inaugural implementation of LLMs' fusion with spatial computing and dynamic reasoning on graph databases containing external geographic data.
    \item  We introduce the IndustryScopeQA benchmark to evaluate the IndustryScopeGPT framework's performance. Experiments on site recommendation and industrial park planning confirm that the IndustryScopeKG dataset and framework enhance the efficiency and adaptability of LLMs. 
\end{itemize}

\begin{table*}[]
\caption{Comparison of IndustryScopeKG with Other Urban Datasets}
\label{tab:dataset}
\begin{threeparttable}[b]
\begin{tabular}{cccccccc}
\toprule
Dataset &
  \begin{tabular}[c]{@{}c@{}}Image\\ (Street View)\end{tabular} &
  \begin{tabular}[c]{@{}c@{}}Socioeconomic\\ Indicators\end{tabular} &
  \begin{tabular}[c]{@{}c@{}}Geographical\\ Data\end{tabular} &
  \begin{tabular}[c]{@{}c@{}}Semantic\\ Feature\end{tabular} &
   Multi-scale &
  Size &
  \begin{tabular}[c]{@{}c@{}}Open\\ Source\end{tabular} \\
  \midrule
WANT~\cite{Want}     &  & \checkmark  & \checkmark &  &  & $100^*$ &  \\
$O^2$-SiteRec~\cite{O2}     &  &  & \checkmark &  &  & 39,465 &  \\
UrbanKG~\cite{UrbanKG}     & \checkmark & \checkmark & \checkmark &  &  &    17,407,159&  \\
UrbanKGent~\cite{UrbanKGent}  &  & \checkmark & \checkmark &  &  & 67,978 & \checkmark \\
OpensiteRec~\cite{OpenSiteRec} &  & \checkmark & \checkmark &  &  & 6,170,925*  & \checkmark \\
KnowSite~\cite{KnowSite} &  & \checkmark & \checkmark & \checkmark &  & 920,504 & \checkmark  \\
UUKG~\cite{UUKG}        &  &  & \checkmark &  & \checkmark & 1,490,680  & \checkmark  \\
\textbf{ IndustryScopeKG(Ours)}        & \checkmark & \checkmark & \checkmark & \checkmark & \checkmark &  51,684,939 & \checkmark  \\
\bottomrule
\end{tabular}
\begin{tablenotes}
    \item * denotes the approximate number w.r.t.the corresponding paper
    \item  \textit{For tabular data, `size' counts the number of data points, and for KG, it counts the number of triples in total.}

\end{tablenotes}
\end{threeparttable}
\end{table*}

\section{Related Work}

\subsection{Urban  Intelligence and Dataset}

Researchers utilize deep learning models to extract representations from urban data like satellite images, points of interest (POI), and road networks~\cite{POI_representation,human,beyond}, but their lack of interpretability and generality across different tasks limits applications in urban settings. ReCo~\cite{Reco} provides dataset with precise coordinates for residential layout planning. However, the reliance on single data source restricts the integration of multi-modal data, impacting complex urban analysis. Initiatives like OpenSiteRec~\cite{OpenSiteRec} and UUKG~\cite{UUKG} employ heterogeneous graphs for brand site recommendations and spatio-temporal predictions. These urban knowledge graphs organize urban entities comprehensively but struggle with limited, task-specific datasets and a scarcity of public resources~\cite{UrbanKGent}. For a comparison of our dataset with other urban dataset, see Table ~\ref{tab:dataset}.

\subsection{LLMs Reasoning}

LLMs exhibit strong capabilities through enhanced reasoning abilities via logical structures like chains~\cite{CoT}, trees~\cite{ToT}, and graphs~\cite{ToG}. Recent advancements enable large models to access internal and external knowledge for improved decision-making~\cite{Rag_survey}. LLMs are increasingly used as central controllers for autonomous agents with human-like decision-making skills~\cite{Agent_survey}. They have spurred innovation in urban research, such as developing mobility strategies~\cite{urban_residents}, simulating disease spread~\cite{Epidemic}, urban planning~\cite{Urban_Planning_Li} and complex spatio-temporal question answering~\cite{UrbanGPT}. However, these methods primarily focus on textual spatio-temporal features, neglecting multi-modal urban knowledge.

\section{IndustryScopeKG}

\subsection{Data Collection and Pre-processing}

\subsubsection{Data Acquisition}

We opted to acquire multi-source spatio-temporal data from the Shanghai, China, considering the richness and availability of information relevant to industrial parks. The data sources mainly come from three aspects:

\begin{figure}[h]
	\centering
	\includegraphics[width=0.4\textwidth]{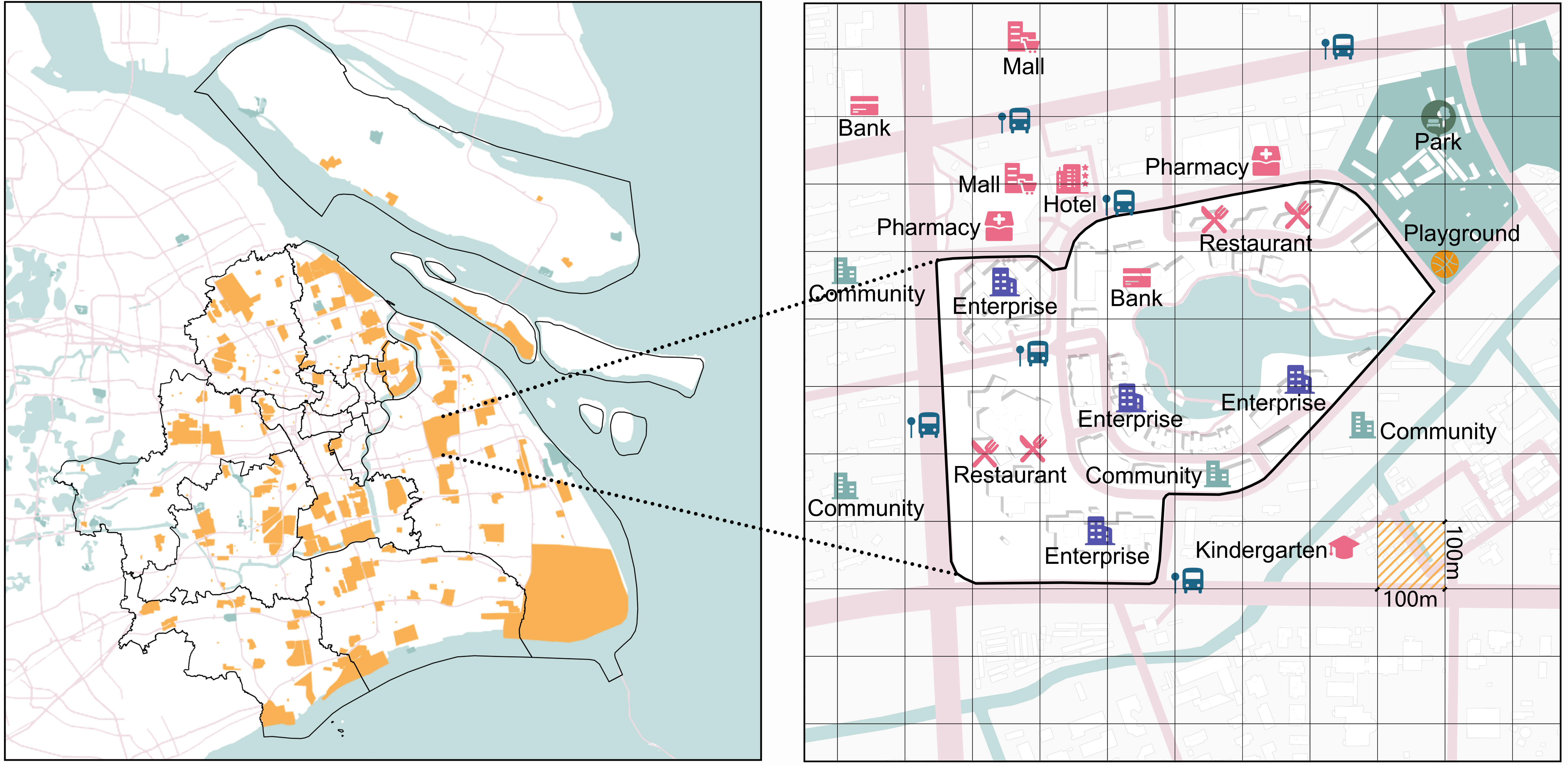}
	\caption{Park vectorization and grid processing.}
	\label{fig2}
\end{figure}

{\bfseries Urban Geospatial Data:} Includes (1) building footprints, areas of interest (AOI), and mobility data based on mobile positioning from Baidu Map; (2) POI and transport station data from Amap.

{\bfseries Corporate Data:} Includes (1) industrial and commercial enterprise registration data from Qichacha; (2) patent and software copyright data from the National Intellectual Property Administration; (3) data on listed companies, state-owned, high-tech, and small and medium-sized enterprises (SMEs) in technology, along with overseas cooperative company data from Macrodatas.

{\bfseries Socioeconomic Data:} Includes (1) census data and regional GDP data from government reports; (2) housing prices from Beike.
All data is collected from open sources to meet ethical regulations. 

\subsubsection{Data Pre-processing}

Extensive data preprocessing was necessary, focusing on geospatial tasks such as park vectorization, grid processing and spatial positioning, as illustrated in Figure~\ref{fig2}. We manually outlined the vector boundaries of each industrial park using maps of Shanghai's industrial parks, converting text addresses into geocoded latitude and longitude coordinates. Additionally, we standardized the coordinate systems of the multi-source data to the Baidu system for accurate positioning within industrial park boundaries. Integrating visual models, we employed semantic segmentation of street view images to calculate visual metrics ~\cite{View_index}, used object detection to assess permeability, and trained models on street view charm value based on expert ratings.

\begin{figure*}[htbp]
	\centering
	\includegraphics[width=0.9 \textwidth]{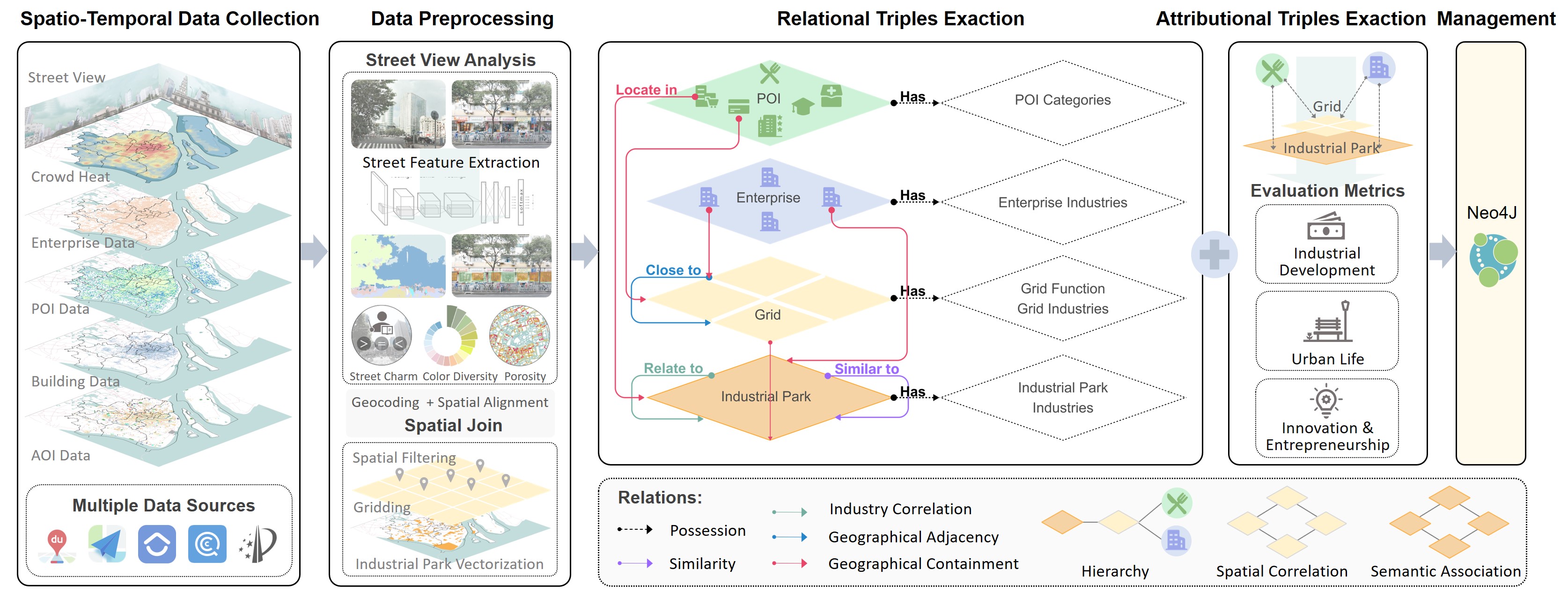}
	\caption{ IndustryScopeKG construction pipeline. It follows the process from data collection and preprocessing to triple extraction and management, forming a graph with hierarchy, spatial correlation, and semantic association.}
	\label{fig3}
\end{figure*}

\subsection{Knowledge Graph Construction}

\subsubsection{IndustryScopeKG}

\textbf{Definition.} We define the graph \( G = (E, S, Y) \), where \( E = \{e_1, e_2, \ldots, e_{|E|}\} \) is a set of \( |E| \) entities, \( S \) denotes the set of relational triples, and \( Y \) encompasses the set of attributional triples. Specifically, \textbf{(1) Relational Triples:} \( S \subseteq E \times R \times E \) represents a collection of triples that delineate relationships between entities, with \( R \) constituting a set of \( |R| \) distinct relations. For instance, “Company -- Located In -- Industrial Park”;  \textbf{(2) Attributional Triples:} \( Y \subseteq E \times A \times V \) constitutes triples indicating attributes of entities, where \( A = \{A_1, A_2, \ldots, A_{|A|}\} \) represents a collection of \( |A| \) attributes, with each attribute \( A_i \in A \) paired with a corresponding set of values \( V_i \in V \). For example, “Industrial Park -- Number of Companies -- 500”. IndustryScopeKG was built following our construction pipeline, as shown in Figure~\ref{fig3}.

\subsubsection{Relational Triples Extraction}

\textbf{Entity Extraction.} For the IndustryScopeKG, we extract entities from 8 major categories and 32 sub-categories. The major categories include: \textbf{(1) Industrial Parks,} encompassing 264 industrial parks in Shanghai. \textbf{(2) Grids,} which are 128,866 fine-grained spatial grids derived from gridding industrial parks. \textbf{(3) Grid Dominant Functions,} identified from calculating the dominant POI within grids. It is then adjusted using AOI, resulting in 15 types of grid functions, e.g., business offices, commercial services, residential areas, among others. \textbf{(4) POI,} serving as the basic functional units and places, including 15 POI sub-categories such as residential, green spaces, business offices, commercial services, etc. \textbf{(5) Enterprises,} including entities of 1,058,656 enterprises within parks. \textbf{(6) Enterprise Industries,} comprising primary, secondary, tertiary industries, and scope of operations, divided into 4 sub-categories. \textbf{(7) Industrial Park Industries,} covering planned industries, leading primary, secondary, and tertiary industries, and scope of operations, divided into 5 sub-categories. These are established based on frequency and significance of their occurrence within enterprises in parks. For industrial park \( P \), with enterprises set \( E_P \), and industry categories \( C_{1,P}, C_{2,P}, C_{3,P} \) corresponding to primary, secondary, and tertiary sectors, the leading industry and scope of operation are identified as:
\begin{equation}
    L_{k,P} = \underset{c_{k} \in C_{k,P}}{\mathrm{argmax}} \left( \frac{|\{e \in E_P | e \text{ is categorized as } c_k\}|}{|E_P|} \right)
\end{equation}

\begin{equation}
O_{s,P} = \underset{s \in S_P}{\mathrm{argmax}} \left( \frac{|\{e \in E_P | \text{scope } s \text{ is listed in } e's \text{ operations}\}|}{|E_P|} \right)
\end{equation}

where \( L_{k,P} \) is the leading industry for category level \( k \) and \( O_{s,P} \) represents the main scope of operations based on enterprise activities in \( P \). \textbf{(8) Grid Industries,} following a similar classification structure to industrial parks, are divided into 4 sub-categories. 

\textbf{Relation Extraction.} We extracted spatial and semantic relationships. Spatial relationships include geographical containment and adjacency; semantic relationships cover similarity, industry correlations between industrial parks, and possession.

\textbf{Geographical Containment:} This explains how one entity is located within another entity, categorized into three types: (1) POI/Enterprise Located in Grid, (2) POI/Enterprise Located in Industrial Park, and (3) Grid Located in Industrial Park.

\textbf{Geographical Adjacency:} This refers to spatial proximity between entities, detailed in two types: (1) Grid Adjacent to Grid, indicating adjacency between grids; (2) Industrial Park Adjacent to Industrial Park, specifying proximity between parks.

\textbf{Similarity:} Indicates the resemblance between industrial parks, expressed as Industrial Park Similar to Industrial Park. We compute the embeddings for each industrial park's unique features and assess park similarities based on cosine distance. 

\textbf{Industry Correlation:} Denotes the connection between industrial parks based on industry characteristics, expressed as Industrial Park Related to Industrial Park. We derive embeddings for each park's industry-related aspects, such as planned industries and leading industries. An industry correlation link is forged between two parks if their industry similarity surpasses a threshold of 0.9.

\textbf{Possession:} Connects entities with what they possess. For example, Industrial Park Has Planned Industries, Grid Has Leading Primary Industries, and Enterprise Has Scope of Operations.

\begin{table*}[]
\caption{The Statistics of Entities in IndustryScopeKG}
\label{tab:industrial_data_overview2}
\begin{tabular}{@{}ccccccc@{}}
\toprule
Basic Statistics &
  Industrial Park &
  Grid &
  Grid Dominant Function &
  POI &
  Enterprise &
  Total \\ \midrule
   Count &
  264 &
  128,866 &
  15 &
  112,931 &
  1,058,656 &
  1,300,732 \\ \midrule
(Leading) Industries &
  Primary &
  Secondary &
  Tertiary &
  Scope of Operations &
  Planned &
   \\ \cmidrule(r){1-7}
\begin{tabular}[c]{@{}c@{}}Industrial Park \end{tabular} &
  202 &
  258 &
  261 &
  261 &
  70 &
  1,052 \\
\begin{tabular}[c]{@{}c@{}}Grid \end{tabular} &
  1,142 &
  6,270 &
  10,281 &
  20,246 &
  / &
  37,939 \\
\begin{tabular}[c]{@{}c@{}}Enterprise\end{tabular} &
  18 &
  90 &
  392 &
  891,814 &
  / &
  892,314 \\ \bottomrule
\end{tabular}
\end{table*}

\begin{table}[htbp]
\caption{The Statistics of Triples in IndustryScopeKG}
\label{tab:industrial_data_overview1}
\scalebox{0.9}{
\begin{tabular}{@{}ccc@{}}
\toprule
Relation &
  Head \& Tail Entity &
  \begin{tabular}[c]{@{}c@{}}Triple\\  Records\end{tabular} \\ \midrule
Locate in &
  \begin{tabular}[c]{@{}c@{}}(POI, Grid)\\ (Enterprise, Grid)\\ (POI, Industrial Park)\\ (Enterprise, Industrial Park)\\ (Grid, Industrial Park)\end{tabular} &
  2,516,160 \\
Adjacent to &
  \begin{tabular}[c]{@{}c@{}}(Grid, Grid)\\ (Industrial Park, Industrial Park)\end{tabular} &
  488,401 \\
Similar to &
  (Industrial Park, Industrial Park) &
  3,765 \\
Related to &
  (Industrial Park, Industrial Park) &
  10,687 \\
Has &
  \begin{tabular}[c]{@{}c@{}}E.g., (Industrial Park, Planned Industries)\\ (Grid, Leading Scope of Operations)\\ (Grid, Dominant Functions)\end{tabular} &
  4,252,341 \\
Attribution &
  {\color[HTML]{333333} \begin{tabular}[c]{@{}c@{}}(Industrial Park, Value) (with 111 attributions)\\ (Grid, Value) (with 82 attribution)\\ (POI, Value) (including 15 attributions)\\ (Enterprise, Value) (with 36 attributions)\end{tabular}} &
  {\color[HTML]{333333} 44,413,585} \\ \bottomrule
\end{tabular}
}
\end{table}

\subsubsection{Attributional Triples Extraction}

To address the challenge of managing diverse entity attributes in a graph database, we use attributional edges for efficient navigation and analysis. Establishing a comprehensive evaluation system is crucial for the robust development of industrial parks, involving an in-depth exploration of urban vitality and industrial park evaluation frameworks~\cite{YeYu}. By analyzing regions like Silicon Valley based on People, Economy, Society, Place, and Governance, detailed data reveals the importance of a quantifiable indicator system covering Industrial Development, Urban Life, and Innovation and Entrepreneurship. 

Specifically, \textbf{(1) Industrial Development:} Includes number of enterprises, large-scale enterprises, average registered capital, state-owned enterprises, listed companies, and industrial agglomeration, match degree of planned industries, working population, GDP, office building area and its proportion, etc. \textbf{(2) Urban Life:} Includes accessibility to residential functions, public services, transportation stations, function diversity, density of public service functions,  function richness, function compatibility, average housing price, work-life balance index, etc. \textbf{(3) Innovation and Entrepreneurship:} Includes number of newly registered enterprises, technology-based SMEs, high-tech enterprises, overseas (cooperative) companies, patents and copyrights, and industrial diversity, accessibility to innovation support functions, population with higher education, density of financial services and research functions, etc.

Within this framework, the industrial park encompasses 74 specific sub-indicators, while the grid is detailed through 48 sub-indicators. A correlative computational strategy allows us to integrate the attributes of smaller spatial units like grids into the analysis of larger entities such as industrial parks. 

\subsubsection{IndustryScopeKG Management}

Following the outlined process, we have constructed a knowledge graph that contains 2,232,037 entities and 51,684,939 triples (for statistical details, see Table ~\ref{tab:industrial_data_overview2},~\ref{tab:industrial_data_overview1}). To manage this expansive scale effectively, we employ the Neo4j graph database for storage, querying, and updates. A key advantage of Neo4j is its spatial capabilities, which greatly enhance our ability to perform spatial computations.

\section{IndustryScopeGPT}

\begin{figure*}[htbp]
	\centering
	\includegraphics[width=\textwidth]{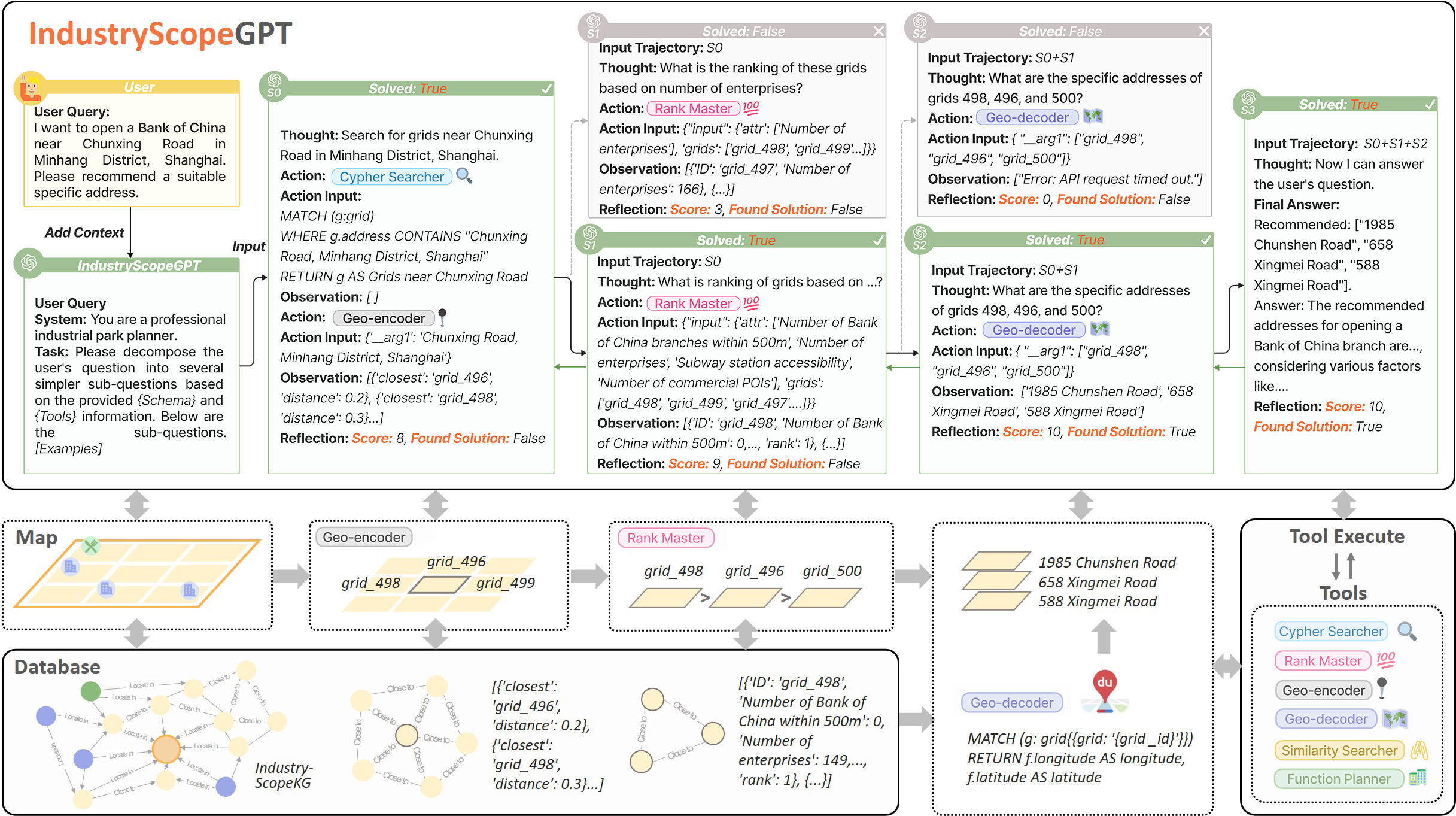}
	\caption{ Overview of IndustryScopeGPT - illustrating conditional financial facility siting based on user queries.}
	\label{fig4}
\end{figure*}

\textbf{Problem Formulation.} 
When addressing user queries, we developed an LLM-driven agent that is capable of generating responses and interacting with external tools that facilitate interaction with graph database. Following action-reflection style work~\cite{LATS}, we define the agent's action state at each step \( t \) as \( a_t \in A \), which is a combination of the text generation \( \hat{A}_t \) and the tool action \( \bar{A}_t \in T \). Such a state pair is represented as \( a_t = (\hat{A}_t, \bar{A}_t) \), where thought \( \hat{A}_t \) is intended to encapsulate an understanding of key information and guide the subsequent action \( \bar{A}_t \). This action is determined by the policy \( \pi(\bar{A}_t \mid Q, a_1, o_1, \ldots, a_{t-1}, o_{t-1}, \hat{A}_t) \). The initial input \( Q \) includes the user's query, task description, schema, tool instructions, and some few-shot examples. To balance exploration and exploitation in finding the best trajectory, we used Monte Carlo Tree Search (MCTS)~\cite{MCTS}. This approach views the large planning space of IPPO-related decision-making tasks as a tree search process. This approach is necessary due to the agent's uncertainty, the improper use of tools or their execution failures, and the potential for better evaluation dimensions or solutions. In this process, each node state is \( S = [Q, a_1, \ldots, a_t, o_1, \ldots, o_t] \). The final answer is derived from the output of the last leaf node on the best trajectory. Each iteration of MCTS consists of four steps. Figure \ref{fig4} provides a typical example of IndustryScopeGPT resolving IPPO processes.

\subsection{Monte Carlo Tree Search Planner}

\textbf{Selection:} The process initiates at the root node (initial state), employing an enhanced UCT (Upper Confidence bounds applied to Trees)~\cite{UCT} algorithm to guide the search towards promising areas for expansion. This approach dynamically balances exploration and exploitation based on aggregated rewards. The core of this refinement is the updated UCT formula: 

\begin{equation}
    UCT(s) = \frac{V(s)}{N(s)} + \omega d^{N(s)} \sqrt{\frac{2 \ln N(p)}{N(s)}}
\end{equation}

Where \( \frac{V(s)}{N(s)} \) represents the average reward of node \( s \), calculated as the total value \( V(s) \) divided by the number of visits \( N(s) \). \( \omega \) represents initial
exploration weight, \( d \) is the decay factor that reduces exploration with each visit, encouraging more exploitation as the node becomes more familiar. \( N(p) \) denotes total visits to the parent node. The search progresses by selecting actions that resolve query or require further exploration until a termination condition is met. It ends when a solution is found, the maximum depth is reached, or if a tool is used more than four times in a row, preventing excessive and irrelevant searches and ensuring meaningful progress towards a solution.

\textbf{Expansion:} During expansion, the search widens by generating new child nodes from feedback on executed actions, all recorded in a long-term memory. Each node undergoes a scalar evaluation to aid future node selection, focusing on simulations that highlight the most promising paths. This phase enables the parallel execution of the best \( k \) potential actions, thereby expanding the exploration domain and enhancing the decision tree's coverage.

\textbf{Reflection:} Post-action execution, outcomes are assessed, incorporating LLM based self-reflection and external feedback to score decisions. To avoid potential misinterpretations due to a lack of context--limited visibility to previous node information--the reflection process also leverages the trajectory's memory. This includes a history of executed actions and generated outcomes.

\textbf{Back-propagation:} Node values along explored paths are updated based on simulations, integrating insights into decision-making to ensure gradual progress towards optimal solutions. This phase employs a recursive update mechanism,
\begin{equation}
    V'(s) = V(s) + \frac{R - V(s)}{N(s)}
\end{equation}

where \( V(s) \) and \( V'(s) \) represent the node \( s \)'s value before and after the update, \( R \) is the reward from simulation, incrementally enhancing tree's accuracy and strategic depth each iteration.

\begin{table*}
  \centering
  \begin{minipage}{0.49\linewidth}
    \caption{Park-Level Site Recommendation}
    \label{tab:1}
    \setlength{\tabcolsep}{3.2pt}
    \begin{tabular}{lccccc}
      \toprule
      Methods and Knowledge & Accuracy & Precision & Recall & F1\\
      \midrule
      GPT-4 w Table/ SE & / & / & / & /\\
      GPT-4 w Cypher Searcher & 0.04 & 0.242 & 0.220 & 0.224\\
      CoT w Tools & 0.127 & 0.319 & 0.319 & 0.320\\
      ReAct w Tools & 0.088 & 0.276 & 0.276 & 0.276\\
      \textbf{IndustryScopeGPT} w Tools & \textbf{0.204} & \textbf{0.443} & \textbf{0.440} & \textbf{0.441}\\
      \bottomrule
    \end{tabular}
  \end{minipage}%
  \hfill
  \begin{minipage}{0.49\linewidth}
    \caption{Conditional Park-Level Site Recommendation}
    \label{tab:2}
    \setlength{\tabcolsep}{3.2pt}
    \begin{tabular}{lccccc}
      \toprule
      Methods and Knowledge & Accuracy & Precision & Recall & F1\\
      \midrule
      GPT-4 w Table/ SE & / & / & / & /\\
      GPT-4 w Cypher Searcher & 0.194 & 0.555 & 0.429 & 0.464\\
      CoT w Tools & 0.166 & 0.488 & 0.488 & 0.488\\
      ReAct w Tools & 0.205 & 0.539 & 0.462 & 0.485\\
      \textbf{IndustryScopeGPT} w Tools & \textbf{0.233} & \textbf{0.659} & \textbf{0.566} & \textbf{0.590}\\
      \bottomrule
    \end{tabular}
  \end{minipage}
\end{table*}

\begin{table*}
  \centering
  \begin{minipage}{0.49\linewidth}
    \caption{Grid-Level Site Recommendation}
    \label{tab:3}
    \setlength{\tabcolsep}{3.2pt}
    \begin{tabular}{lccccc}
      \toprule
      Methods and Knowledge & Accuracy & Precision & Recall & F1\\
      \midrule
      GPT-4 w Table/ SE & / & / & / & /\\
      GPT-4 w Cypher Searcher & 0.064 & 0.368 & 0.368 & 0.368\\
      CoT w Tools & 0.236 & 0.488 & 0.480 & 0.483\\
      ReAct w Tools & 0.232 & 0.509 & 0.428 & 0.457\\
      \textbf{IndustryScopeGPT} w Tools & \textbf{0.292} & \textbf{0.584} & \textbf{0.572} & \textbf{0.577}\\
      \bottomrule
    \end{tabular}
  \end{minipage}%
  \hfill
  \begin{minipage}{0.49\linewidth}
    \caption{Conditional Grid-Level Site Recommendation}
    \label{tab:4}
    \setlength{\tabcolsep}{3.2pt}
    \begin{tabular}{lccccc}
      \toprule
      Methods and Knowledge & Accuracy & Precision & Recall & F1\\
      \midrule
      GPT-4 w Table/ SE & / & / & / & /\\
      GPT-4 w Cypher Searcher & 0.161 & 0.475 & 0.437 & 0.450\\
      CoT w Tools & 0.127 & \textbf{0.550} & 0.395 & 0.433\\
      ReAct w Tools & 0.06 & 0.258 & 0.181 & 0.204\\
      \textbf{IndustryScopeGPT} w Tools & \textbf{0.194} & 0.492 & \textbf{0.483} & \textbf{0.487}\\
      \bottomrule
    \end{tabular}  
  \end{minipage}
\begin{tablenotes}
    \item  \textit{/ represents mostly zero or near-zero metrics.}
\end{tablenotes}    
\end{table*}

\subsection{Decision Support Tools}
\begin{figure}[htbp]
	\centering
	\includegraphics[width=0.476\textwidth]{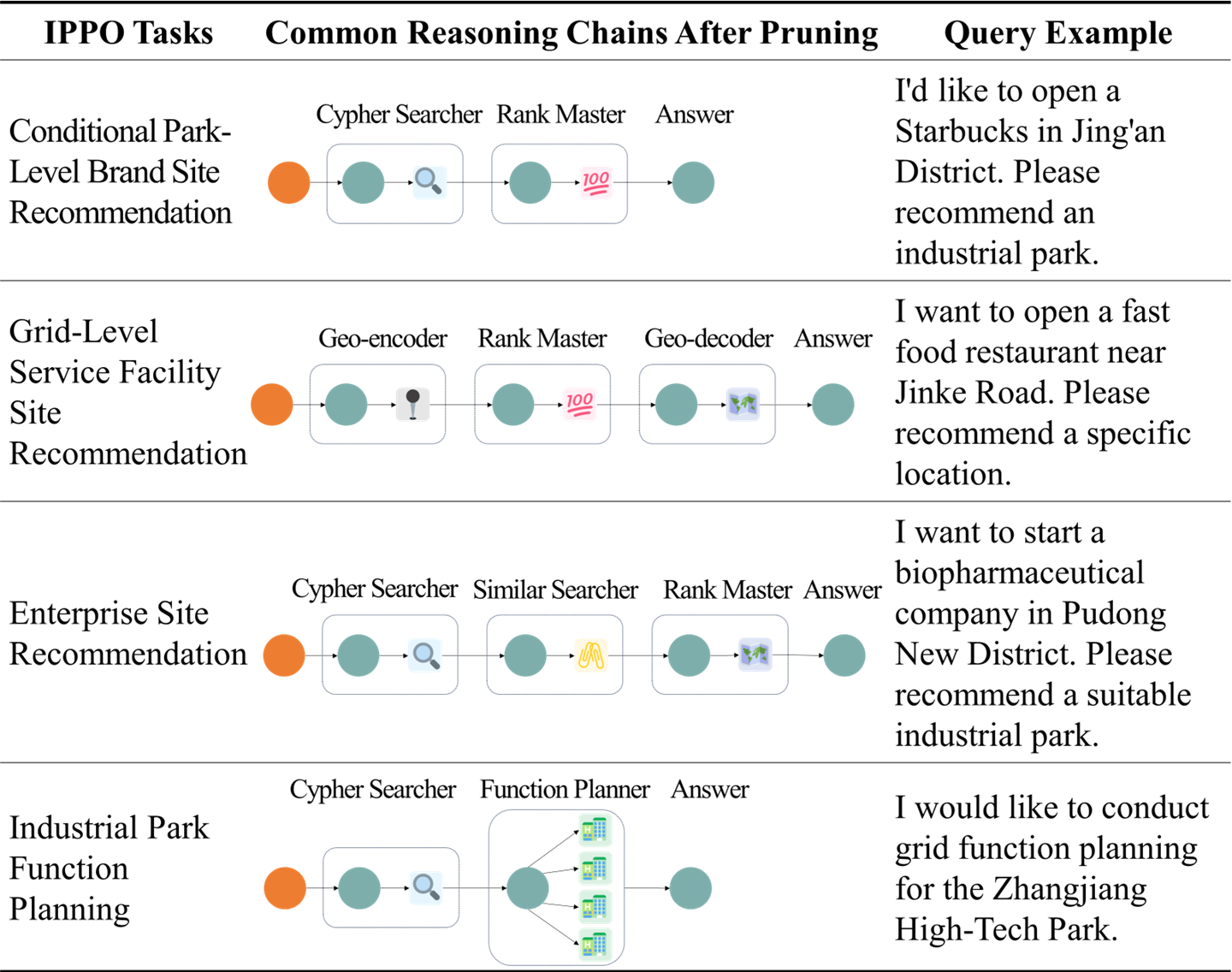}
	\caption{Designed tools and their typical reasoning chains.}
	\label{fig5}
\end{figure}

In crafting a multifaceted decision support system for IPPO, we've developed a suite of nuanced sub-task tools, as depicted in Figure~\ref{fig5}. These tools interact with a Neo4j graph database to facilitate sophisticated queries, analyses, and recommendations, directly addressing the complex needs of urban planners, investors, and businesses.

\textbf{Cypher Searcher:} Tailored to generate and execute Cypher queries based on user inquiries, it sifts through graph database to provide detailed insights on industrial park attributes, facilities, and demographics, streamlining the data retrieval process.

\textbf{Similarity Searcher:} Based on semantic similarity, this tool searches and recommends parks with high similarity to the user's input, e.g., type of business. By analyzing the match between park features and user's specified business or operational criteria.

\textbf{Geo-encoder:} It is designed to convert textual addresses into precise geographic locations, specifically identifying relevant grid IDs in graph database. This conversion allows for a seamless mapping of user-provided locations to the spatial framework of database.

\textbf{Geo-decoder:} Conversely, the tool translates geographic grid IDs, obtained from spatial queries or system recommendations, back into human-readable textual addresses. 

\textbf{Rank Master:} This tool integrates a non-parametric Borda Count rank aggregation method with LLM to rank parks or grids based on selected metrics like accessibility, POI density, and demographics. Formally, for potential sites $\mathcal{C} = {s_1, s_2, \ldots, s_m}$ and evaluation criteria $\mathcal{V} = {v_1, v_2, \ldots, v_n}$, each site $s_i$ is assigned a Borda Count $B(s_i) = \sum_{j=1}^n (m - r_{ij})$, where $r_{ij}$ is $s_i$'s rank for criterion $v_j$.

\textbf{Function Planner:} An assistant designed to propose functional planning suggestions for specified areas. Given a grid, it not only retrieves information for the designated and adjacent grids but also synthesizes these insights to offer strategic planning advice. 

\section{Experiment}

To validate effectiveness of IndustryScopeKG and IndustryScopeGPT framework, we aim to address the following research questions:

\textbf{RQ1:} Can the graph data provided by IndustryScopeKG effectively impact the performance of LLMs on IPPO?

\textbf{RQ2:} How does IndustryScopeGPT perform on tasks based on IndustryScopeKG compared to existing LLM prompting paradigms?

\textbf{RQ3:} How can IndustryScopeGPT aid park functional planning? 

To explore these questions, we designed two typical IPPO tasks.

\subsection{Multi-spatial Scale Facility Siting Recommendation (RQ1, RQ2)}

To improve facility siting flexibility, we introduce a multi-spatial scale recommendation task that leverages LLM's reasoning abilities to analyze and filter spatial attribute features for optimal facility siting, regardless of scale or facility type.

\textbf{Dataset.} We created 20,000 siting questions across various spatial scales and facility categories, incorporating attribute data from industrial parks and grids in the IndustryScopeKG. LLM identified 5-8 key evaluation attributes for each question, resulting in three sets. Domain experts selected one set through consensus-building, and top areas were determined using an optimal ranking method, forming question-answer pairs. We used 200 question-answer pairs covering diverse spatial scopes and facility types for test.

\textbf{Baseline Methods (RQ1).} To assess the effectiveness of IndustryScopeKG, we evaluated the performance of GPT-4 in various configurations: alone, with tabular input for structured data processing, integrated with a search engine to enhance data retrieval capabilities, and combined with Cypher for querying graph databases. This approach allowed us to compare the impact of different data integration methods on the model's ability to handle IPPO tasks.

\textbf{Baseline Methods (RQ2).} For tool invocation, we contrasted IndustryScopeGPT with methods highlighting LLMs' external tool and graph database interaction capabilities. We compared classic prompting with CoT and ReAct methodologies, examining performance across varied siting tasks and spatial scales.

\textbf{Experiment Settings.} The version of gpt-4-0125-preview was used across the experiments, with Cypher Searcher temperature set to 0 and QA model at 0.7. MCTS was calibrated to expand 2 child nodes at each level, with 5 maximum depth, 50 recursion limit.

\textbf{Metrics.} We assessed whether the predicted locations matched those in dataset, measuring Accuracy, Precision, Recall and F1 score.

\textbf{Results.} The evaluation demonstrated that GPT-4, whether operating independently or with tabulated data (where IndustryScopeKG data was structured into tables and embedded into a vector library for retrieval-augmented generation) or search engine, performed poorly in complex site recommendation tasks (RQ1). This highlights the significance of our dataset and graph structure. In contrast, IndustryScopeGPT exhibited superior performance (RQ2), significantly outperforming methods such as those using a Cypher searcher, CoT, and ReAct across most metrics (details in Table ~\ref{tab:1} - ~\ref{tab:4}). For example, in the conditional park-level site recommendation, IndustryScopeGPT demonstrated its capability to effectively leverage structured industrial data, achieving a precision of 0.659, and F1 score of 0.590 (Table ~\ref{tab:2}). This showcases its superior performance in optimizing decision-making for site recommendation tasks.

\subsection{Industrial Park Functional Planning (RQ3)}
This task focuses on the optimization of various grid functions within an industrial park, aiming to fulfill the foundational needs and balanced layout of functionalities. The planning process involves the examination of each grid within a specific industrial park, extracting information from targeted and adjacent grids and integrating these attribute insights to assign a function to each grid.

\textbf{Baseline Methods (RQ3).} Traditional models such as LightGBM~\cite{LightGBM} and GCN~\cite{GCN} were used for comparison to evaluate IndustryScopeGPT’s performance in functional planning.

\textbf{Dataset.} For training LightGBM and GCN, datasets were prepared matching the input formats required by these models. All 128,866 grid attributes were used as features. For GCN, grid adjacency relationships were used to define edges, and grid functions pre-calculated in  IndustryScopeKG, corresponding to 15 categories, served as labels. The data was split into 7:3 for training and testing.

\textbf{Metrics.} Given the absence of uniform standards for planning evaluation, we focused on improvements to the existing functional layout. We utilized the Hill numbers formula~\cite{Hill} to quantify functional diversity, which is defined as:

\begin{equation}
    H_q = \left( \sum_{i=1}^S p_i^q \right)^{\frac{1}{1-q}}
\end{equation}

where \( p_i \) represents the proportion of the \( i \)-th function type in park, and \( q \) determines the emphasis on abundance. Specifically, \( q = 0 \) quantifies the Richness, or the total count of distinct functional types; \( q = 1 \) corresponds to Shannon Entropy, which accounts for the proportion of each function; \( q = 2 \) is tied to Simpson's Index, focusing on the prevalence of dominant functions. These measures are vital for evaluating space utilization and ensuring integration of diverse functions to support industrial activities.

\begin{table}[]
\caption{Detailed Analysis of Grid Functions by Methods}
\label{tab:grid_functions_analysis}
\scalebox{0.765}{
\begin{tabular}{@{}c|p{6cm}|c|c|c@{}}
\toprule
Methods & Grid Functions & q=0 & q=1 & q=2 \\ \midrule
\textbf{Real Situation} & 
\textbf{Park A}: Green Space & 1 & 1 & 1 \\
\textbf{} & \textbf{Park B}:
 Residential: Sports Recreation: Green Space = 8:3:1 & 3 & 2.39 & 2.11 \\
\textbf{} & \textbf{Park C}: Sports Recreation: Green Space: Residential = 12:5:1 & 2 & 1.81 & 1.67 \\
\textbf{LightGBM} & \textbf{Park A}: 
Residential: Green Space = 1:6 & 2 & 1.51 & 1.32 \\
\textbf{} & \textbf{Park B}: Residential: Sports Recreation: Green Space = 8:3:1 & 3 & 2.28 & 1.95 \\
\textbf{} & \textbf{Park C}: Sports Recreation: Green Space: Residential = 12:5:1 & 2 & 1.81 & 1.67 \\
\textbf{GCN} & 
\textbf{Park A}: Residential: Green Space = 3:4 & 2 & 1.98 & 1.96 \\
\textbf{} & \textbf{Park B}: Residential: Green Space = 1:3 & 2 & 1.76 & 1.60 \\
\textbf{} & \textbf{Park C}: Residential: Green Space = 4:5 & 2 & 1.99 & 1.98 \\
\textbf{IndustryScopeGPT} & 
\textbf{Park A}: Healthcare: Commercial Services: Business Office: Residential: Education: Culture: Life Services: Research Institutions = 1:1:2:1:4:1:2:2 & 8 & 7 & 6.13 \\
\textbf{} & \textbf{Park B}: Education: Commercial Services: Green Space: Residential: Business Office: Sports Recreation: Life Services: Dining Services: Healthcare: Research Institutions: Traffic = 8:6:5:4:3:3:3:1:1:1:1 & 11 & 8.76 & 7.54 \\
\textbf{} & \textbf{Park C}: Education: Green Space: Business Office: Residential: Commercial Services: Sports Recreation: Life Services = 6:4:2:2:2:1:1 & 7 & 6.22 & 5.59 \\
\bottomrule
\end{tabular}
}
\end{table}

\begin{figure}[htbp]
	\centering
	\includegraphics[width=0.48\textwidth]{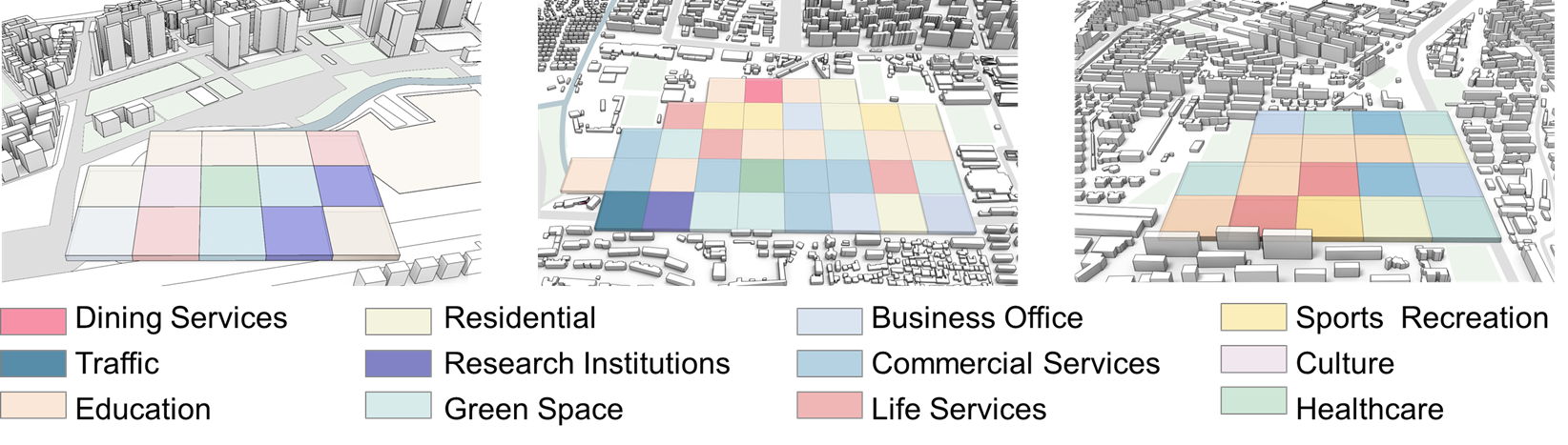}
    	\caption{Utilizing IndustryScopeGPT for grid function planning in parks (From left to right: Park A, Park B, Park C). IndustryScopeGPT achieved optimal functional diversity.}
	\label{fig6}
\end{figure}

\textbf{Case study.} We focused on three distinct parks: Zhangjiang Artificial Intelligence Island (Park A), Dachang Urban Industrial Park (Park B), and Xinyefang Global Sci-Tech Innovation District (Park C) (Figure~\ref{fig6}). Each park was chosen for its unique characteristics and specific challenges it presents in terms of spatial planning. IndustryScopeGPT was tested against two classic models, as well as the real-world scenarios. The results showed that IndustryScopeGPT significantly outperformed these models in the metrics across the three parks. For example, in Park A, known for its focus on green space, IndustryScopeGPT was able to ensure a better balance between green areas and industrial development. In Park B, which combines residential, sports, and green areas, IndustryScopeGPT demonstrated superior ability to integrate diverse functions into a cohesive plan that supports both living and recreational activities. Similarly, in Park C, IndustryScopeGPT promotied an effective synergy between education and commercial activities (see Table ~\ref{tab:grid_functions_analysis}).

\section{Conclusion}

This study presents a transformative approach to intelligent planning and operation of industrial parks by integrating LLMs with the large-scale multi-modal, multi-level IndustryScopeKG. The IndustryScopeGPT framework, using advanced retrieval and reasoning strategies like Monte Carlo Tree Search, sets new standards in IPPO tasks, enhancing adaptability and interpretability. Empirical results confirm significant improvements in site recommendation and functional planning. This initiative underscores IndustryScopeKG's crucial role in advancing urban industrial applications and future LLMs integration into urban development strategies.

\textbf{Limitations and future work.} IndustryScopeKG currently focuses on a single city and plans are in place to expand to more cities. Additionally, the framework's reliance on prompt token consumption and computational resources may present challenges. Tree-structured search may consume significantly more tokens compared to simpler prompting methods. Future improvements will focus on enhancing efficiency and reducing these costs.

\begin{acks}
This work was supported by the National Natural Science Foundation of China (U23B2057),  the NSFC Young Scientists Fund (No. 62006203), the Top Discipline Plan of Shanghai Universities-Class I (No. 2022-3-YB-02), the Science and Technology Commission of Shanghai Municipality, the Innovation Design and Intelligent Manufacturing Discipline Cluster Project at Tongji University, the Research Grants Council of the Hong Kong Special Administrative Region (No. PolyU/25200821), the Innovation and Technology Fund (No. PRP/047/22FX), PolyU Research Centre on Data Science and Artificial Intelligence (No. 1-CE1E), and PolyU Embodied Artificial Intelligence Lab (No. N-ZGNN).
\end{acks}


\bibliographystyle{ACM-Reference-Format}
\bibliography{sample-base}

\appendix

\end{document}